# T2D-Bench: Evidence-Gated Evaluation of LLM Outputs for Type 2 Diabetes Using a Multi-Layer Clinical–Lifestyle Knowledge Graph

**Saba A. Farahani, MS[1], Hung Cao, PhD[1], Ramesh Jain, PhD[1], Amir M. Rahmani, PhD[1]**
**[1]University of California, Irvine, Irvine, CA, USA**

**Abstract**
*Large language models (LLMs) can produce clinically fluent recommendations for type 2 diabetes while failing to satisfy guideline constraints or explicitly justify lifestyle-related glycemic claims. We present T2D-Bench, a reproducible benchmark and evidence-gated evaluation framework for testing whether LLM outputs satisfy explicit, graph-checkable evidence requirements. T2D-Bench is built on a multi-layer clinical–lifestyle knowledge graph that combines a biomedical spine (UMLS, DrugBank, SIDER), computable ADA Standards of Care rules, and lifestyle knowledge connected through a mechanistic bridge to glycemic laboratory effects. Across 100 structured vignettes spanning diagnosis, medication safety, and adversarial lifestyle conflicts, baseline outputs failed benchmark-defined evidence-path checks in 35% of cases for GPT-4o-mini and 33% for GPT-4o. The evidence gate detects unsupported omissions and uses constrained revision to bring outputs into verifier-level compliance with benchmark-defined evidence requirements. These results show that computable evidence constraints can make unsupported clinical omissions explicit, measurable, and correctable in diabetes-focused LLM outputs.*

**Code and Data Availability:** Code to build the knowledge graph, benchmark vignettes, and evaluation scripts are available at https://github.com/Saba-Farahani/t2d-bench-.git.

## Introduction

Large language models (LLMs) can generate fluent clinical explanations and recommendations, making them attractive for scalable decision support in chronic disease management. Prior studies suggest that LLMs encode substantial clinical knowledge and can perform well on medical question-answering tasks, yet reliability remains a major concern in safety-critical settings [1,2]. Even plausible outputs may violate deterministic clinical constraints, overlook contraindications, or present confident explanations without verifiable support. In healthcare, such failures are especially consequential because they can appear clinically reasonable while lacking the evidence needed for safe use [3,4].

A central limitation of many current clinical LLM evaluations is their emphasis on lexical similarity, overall plausibility, or subjective expert judgment [2,5]. Such approaches may fail to test whether a model's claims are explicitly supported by authoritative evidence or whether it can handle multi-factor patient scenarios without silently inventing missing mechanistic steps [6]. In type 2 diabetes (T2D), safe decision-support outputs often require combining biomedical entities such as diagnoses, laboratory values, and medications with guideline logic and lifestyle context, including diet, physical activity, and sleep, all of which can materially influence glycemic outcomes [7].

To address this gap, we introduce T2D-Bench, a reproducible benchmark and evidence-gated evaluation framework for testing whether diabetes-focused LLM outputs satisfy computable clinical and mechanistic evidence requirements (Figure 1). T2D-Bench is built on a multi-layer knowledge graph designed for deterministic verification. The graph combines (i) a biomedical spine derived from controlled terminologies and drug knowledge resources [8–10]; (ii) computable American Diabetes Association (ADA) Standards of Care rules encoded as rule primitives, including diagnostic thresholds and renal medication constraints [7]; and (iii) a lifestyle layer derived from diet, activity, and sleep knowledge sources, connected through an explicit mechanistic bridge to glycemic laboratory effects [11–13].

This design supports an Evidence Gate that checks whether an LLM output satisfies vignette-specific evidence requirements in the graph and triggers constrained revision or flagging when required support is missing. We evaluate the framework on 100 structured vignettes spanning diagnosis, medication safety, lifestyle effects, and adversarial multi-factor conflicts. Relative to baseline outputs, evidence gating intercepts responses that fail benchmark-defined evidence requirements and brings them into verifier-level compliance through constrained revision.

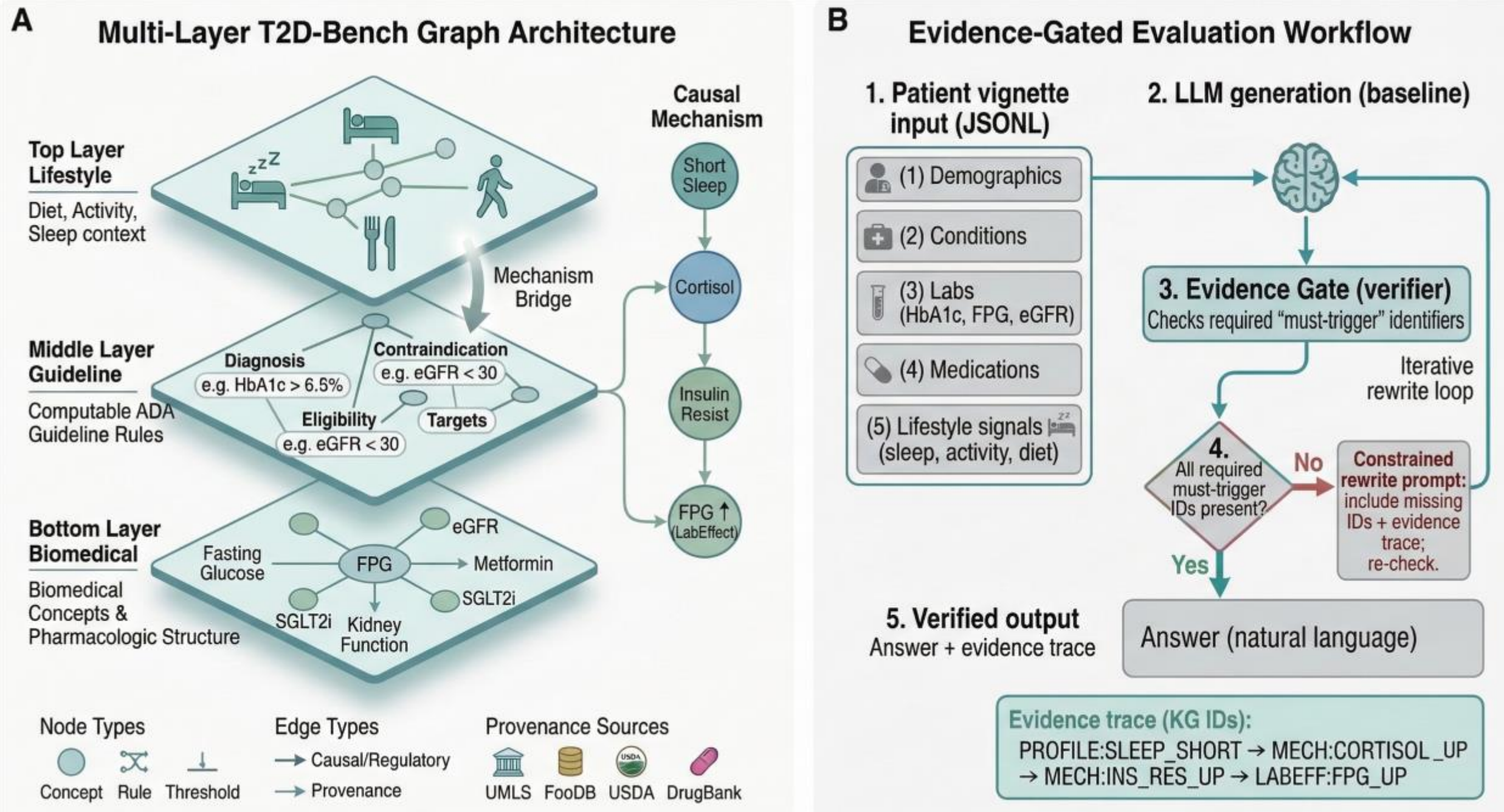

**Figure 1.T2D-Bench overview: multi-layer knowledge graph and evidence-gated evaluation workflow. (A)** The multi-layer T2D knowledge graph comprises a biomedical spine (Layer 1; UMLS and drug resources), computable ADA guideline primitives (Layer 2; rules and thresholds), and lifestyle profiles linked through an explicit mechanistic bridge to glycemic laboratory effects (Layer 3). **(B)** In the evidence-gated workflow, an LLM generates a baseline response, and a deterministic verifier checks whether vignette-specific evidence identifiers are present in the benchmark graph slice, $G_{bench}$. These identifiers include gold must-trigger Rule IDs and mechanistic bridge tokens. If required evidence is missing, a constrained revision is requested, and the response is reverified until it passes or the revision budget is exhausted.

Our central premise is that clinically plausible output should not be treated as evidence-grounded output; T2D-Bench is designed to make hidden omissions explicitly measurable, auditable, and correctable.

**Contributions.** This work makes three contributions. First, we develop a multi-layer, provenance-aware T2D knowledge graph that connects biomedical concepts, computable guideline logic, and lifestyle context through an explicit mechanistic bridge. Second, we introduce T2D-Bench, a benchmark of 100 structured vignettes paired with predefined evidence identifiers for testing whether diabetes-focused LLM outputs satisfy computable clinical and mechanistic evidence requirements. Third, we present an evidence-gated evaluation framework that detects unsupported omissions and uses constrained revision to achieve benchmark-defined verifier compliance.

## Related Work

**LLMs and Clinical Evaluation**. LLMs have shown strong performance on medical question answering and clinical text tasks [1,2], yet their use in clinical decision support remains limited by inconsistent adherence to safety requirements and guideline logic [5]. Hallucination and ungrounded generation are well-established failure modes in natural language generation and are especially consequential in healthcare, where plausible language can obscure unsafe or unsupported recommendations [3,4]. Many current evaluation settings still emphasize surface-form correctness, overall plausibility, or subjective expert scoring, which may fail to detect deterministic errors such as omitted thresholds, missed contraindications, or unsupported mechanistic claims [5,6]. In contrast, our work focuses on whether model outputs satisfy explicit, benchmark-defined evidence requirements rather than whether they merely appear clinically reasonable.

**Biomedical terminologies and drug knowledge resources.** Biomedical terminologies and knowledge graphs provide the structured foundation needed to normalize and connect clinical entities across heterogeneous sources. UMLS remains a central resource for integrating biomedical concepts and identifiers at scale [8]. Drug knowledge resources such as DrugBank and SIDER contribute structured relations for targets, interactions, and adverse effects that are important for medication safety assessment [9,10]. However, these resources are primarily designed for semantic coverage and interoperability rather than for executable decision logic or benchmark-level verification. T2D-Bench builds on these resources by organizing them into a graph structure that supports deterministic evidence checking.

**Guideline logic and computable constraints.** Clinical decision support has long relied on formalizing guideline knowledge into rule-based representations. In diabetes care, the ADA Standards of Care specify threshold-based diagnostic criteria and condition-dependent medication constraints that are well suited to deterministic verification [7]. More recent neuro-symbolic and retrieval-augmented generation approaches often ground LLM outputs by retrieving guideline text, but they do not consistently encode guideline logic as computable constraints that can verify whether an output satisfies required conditions or should be revised when support is missing [6]. Our framework emphasizes this distinction by operationalizing guideline support as explicit, checkable benchmark requirements.

**Lifestyle knowledge and mechanistic support in chronic disease.** Lifestyle factors such as diet, physical activity, and sleep influence glycemic physiology through complex and interacting mechanisms. For example, short sleep has established links to metabolic and endocrine disruption relevant to insulin resistance and glycemic control [13]. Structured resources exist for dietary composition, food composition, and physical activity, including FooDB, USDA databases, and MET compendia [11,12], but these sources are rarely integrated into evaluation settings that require explicit mechanistic support rather than plausible narrative explanation alone. T2D-Bench addresses this gap by linking lifestyle factors to mechanistic bridge tokens and downstream laboratory effects, enabling deterministic evaluation of evidence paths.

## Method

T2D-Bench evaluates whether LLM outputs satisfy explicit, deterministic clinical and mechanistic evidence requirements rather than relying on surface plausibility. The framework consists of three components: (i) a multi-layer clinical–lifestyle knowledge graph, (ii) a benchmark of 100 structured vignettes paired with gold "must-trigger" evidence identifiers, and (iii) an Evidence Gate that verifies outputs and, when necessary, requests constrained revision to satisfy required guideline and mechanistic evidence paths. Figure 1 summarizes the overall architecture and verification loop.

**Multi-layer knowledge graph construction**. We represent domain knowledge as a typed directed property graph, $G = (V, E)$, with globally unique node identifiers and typed edge predicates. Nodes include biomedical entities such as conditions, medications, and laboratory tests; guideline primitives such as rules and thresholds; and lifestyle and mechanistic primitives such as profiles, mechanisms, and laboratory effects. Edges encode drug relations, guideline constraints, and mechanistic links used to operationalize lifestyle-to-laboratory pathways.

*Layer 1 (Biomedical spine).* We construct a diabetes-focused subgraph from UMLS concepts and relations and augment it with pharmacologic structure from DrugBank-derived relations, including drug–drug interactions, drug targets, drug–food interaction statements, and SIDER adverse-effect links. This layer provides normalized identifiers, semantic typing, and pharmacovigilance structure for downstream verification.

*Layer 2 (Computable guideline logic).* This layer encodes selected statements from the ADA Standards of Care (2026) as machine-checkable graph primitives. Each guideline statement is represented as a Rule node linked to one or more Threshold nodes, defined by operator, value, and unit, and to a consequent representing the clinical implication, such as diagnosis, contraindication, or treatment eligibility. This design supports deterministic checking of threshold- and condition-dependent constraints, including HbA1c and fasting plasma glucose diagnostic thresholds, eGFR-dependent metformin contraindications, and SGLT2 inhibitor eligibility in chronic kidney disease.

*Layer 3 (Lifestyle context and mechanistic bridge).* This layer captures lifestyle context and links it to glycemic physiology. We curate lifestyle knowledge from USDA and FooDB for dietary composition, the Compendium of Physical Activities for MET-coded activity, and the Sleep Domain Ontology for sleep-related concepts. To support attribution in multi-factor scenarios, we construct a bridge subgraph that maps LifestyleProfile nodes, such as short sleep or high sugar intake, to Mechanism nodes, such as cortisol elevation or insulin resistance, and then to LabEffect nodes aligned with Layer 2 laboratory identifiers, such as increased fasting glucose or elevated HbA1c. Mechanistic edges are explicitly typed, for example CAUSALLY_INCREASES and CAUSALLY_DECREASES, and stored with provenance metadata including source, evidence tier, and confidence to support auditability.

**Table 1**. Core components of T2D-Bench and their roles in evidence-gated evaluation.

| Component | Description and contents | Role in evaluation framework | Scale |
|---|---|---|---|
| **Integrated multi-layer KG** | Biomedical entities, computable guideline rule primitives, and a lifestyle-to-mechanism bridge connected by typed edges | Full source graph supporting provenance, extensibility, and derivation of $G_{bench}$ | 2,101,124 nodes; 495,161 edges |
| **Benchmark slice ($G_{bench}$)** | Minimal subgraph containing all vignette must-trigger identifiers and required evidence paths | Supports fast, deterministic verification and reproducible evaluation | 15,792 nodes; 102,443 edges |
| **Structured vignettes (JSONL)** | Patient demographics, conditions, laboratory observations, medications, and lifestyle signals | Standardized inputs for both baseline and evidence-gated evaluation | 100 cases |
| **Gold evidence specifications** | Required ADA Rule IDs (Layer 2) and bridge/mechanism tokens (Layer 3) for each vignette | Define evidence-path compliance for automated scoring | 100 specifications |
| **Evidence Gate** | Deterministic verifier with a constrained revision loop (T=0) | Detects missing evidence identifiers and requests revision or flagging | 1 module |

**Benchmark graph slice for reproducible evaluation**. The fully merged source graph contains 2,101,124 nodes and 495,161 edges. To enable fast and deterministic checking during evaluation, we extract a compact benchmark slice, $G_{bench}$, that retains (i) all Layer 2 rule structures and dependencies, (ii) the Layer 3 profiles and mechanisms referenced by the benchmark, and (iii) the minimal Layer 1 neighborhood needed for included medications and safety relations. The released $G_{bench}$ contains 15,792 nodes and 102,443 edges. We validate slice integrity using reachability checks to ensure that vignette observation nodes connect to all required gold must-trigger identifiers. Table 1 summarizes the benchmark artifacts and their scale.

**Benchmark vignettes and gold evidence targets**. We construct 100 structured vignettes in JSONL format. Each vignette specifies patient demographics, conditions, laboratory observations, medications, and lifestyle signals. Each case is paired with a deterministic gold specification comprising:

- Must-trigger rules (Layer 2): guideline Rule IDs required for the vignette.
- Must-trigger bridge tokens (Layer 3): lifestyle profiles, mechanisms, and laboratory effects required to support mechanistic attribution.

The benchmark spans four scenario families: diagnostic threshold compliance, medication safety under comorbidity constraints, single-factor lifestyle effects, and multi-factor lifestyle conflicts designed to test mechanistic misattribution.

**Evidence Gate: deterministic verification and constrained revision.** For each vignette, the LLM first generates an answer. The Evidence Gate then checks whether the output contains the vignette's required gold must-trigger identifiers. If any required identifier is missing, the system issues a constrained revision request asking the model to revise the answer while explicitly incorporating the missing evidence identifiers as an evidence trace. The revised output is then rechecked. This loop terminates when all required identifiers are present or when a fixed revision budget is exhausted.

**Experimental settings**. We evaluate two settings on the same 100-vignette benchmark:

- Baseline outputs: the model responds without graph-based verification.
- Evidence-gated outputs: the model responds, then undergoes verification and constrained revision if required evidence is missing.

**Step A: Vignette (Structured Input)**

**Case ID:** T2D_CONFLICT_008
**Diagnosis:** Type 2 diabetes (T2D)
**Lab:** FPG 128 mg/dL
**Profiles:** Short sleep; High sugar intake
**Question:** Explain causal contributors to fasting glucose; prioritize without hallucinating.

**Step B: Baseline Output (Unverified)**

Synthesized short Case ID T2D_CONFLICT_008 has elevated FPG. High-sugar diets spike glucose. Short sleep negatively impacts metabolism. **Diet reduction should be prioritized.**

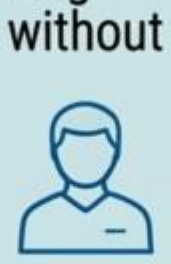

[FAIL] - missing evidence IDs

**Step C: Evidence Gate (ID check)**

**Verification Results:**
Text coherence: PASS
Clinical plausibility: PASS
Evidence IDs: FAIL (missing must-trigger IDs)

**Missing required IDs:**
- MECH:CORTISOL_UP
- MECH:INS_RES_UP
- MECH:PPG_SPIKE
- LABEFF: A1C_UP

**Action:** constrained rewrite requesting missing IDs + evidence trace.

**Step D: Final Output (Verified)**

Synthesized concise **fasting glucose** is 128 mg/dL, consistent with diabetic hyperglycemia.
Chronic **short sleep** leads to elevated **cortisol**, worsening insulin resistance. Prioritize sleep and sugar reduction.

[PASS] - evidence IDs present

**Evidence trace (requiredKG IDs)**
SLEEP_SHORT → CORTISOL_UP → INS_RES_UP → FPG_UP; HIGH_SUGAR → PPG_SPIKE

**Figure 2. Evidence-gated correction of a lifestyle-conflict vignette (T2D_CONFLICT_008). (A)** Structured vignette input with type 2 diabetes, elevated fasting plasma glucose, and conflicting lifestyle signals. **(B)** The baseline output from GPT-4o-mini is clinically plausible but omits required mechanistic evidence identifiers. **(C)** The Evidence Gate detects missing gold must-trigger identifiers and requests constrained revision. **(D)** The revised output includes the required evidence trace and passes automated verification. **Bottom**: gold must-trigger evidence identifiers required for this vignette.

All experiments use deterministic decoding (temperature=0) to support reproducibility. We evaluate two OpenAI models, GPT-4o-mini and GPT-4o, under the same baseline and evidence-gated protocol on the identical 100-vignette benchmark.

**Metrics**. We report three metrics: (i) evidence-path compliance, defined as the fraction of cases whose final output contains all required gold must-trigger identifiers; (ii) intervention rate, defined as the fraction of cases requiring at least one constrained revision; and (iii) scenario-stratified versions of these metrics for each vignette family.

**Results**

**Benchmark-Defined Evidence Compliance**. We evaluated T2D-Bench on 100 structured vignettes using GPT-4o-mini and GPT-4o with deterministic decoding (temperature=0). Evidence-path compliance was defined as whether the model output explicitly contained all vignette-specific required identifiers, including Layer 2 guideline Rule IDs and Layer 3 lifestyle and mechanistic tokens.

In the baseline setting, GPT-4o-mini satisfied all required identifiers in 65/100 cases (0.65), whereas GPT-4o satisfied them in 67/100 cases (0.67). The remaining cases omitted at least one required identifier and therefore failed the benchmark's evidence requirements.

With evidence gating, outputs missing required identifiers were intercepted and revised through constrained revision prompts. After gating, both models achieved 100/100 evidence-path compliance under the automated verifier, with final outputs including an explicit Evidence Trace line listing the required identifiers. This result reflects compliance with benchmark-defined evidence requirements rather than a claim of universal clinical correctness.

**Scenario-Stratified Vulnerability.** Performance differed sharply by scenario family (Table 2). Both models achieved 23/23 compliance on diagnostic threshold vignettes and 23/23 on renal medication-safety vignettes. In contrast, both models showed no baseline compliance on multi-factor lifestyle conflicts (0/30 across the four conflict subtypes), where baseline answers were often clinically plausible but omitted one or more required mechanistic bridge identifiers. Evidence gating corrected these omissions by enforcing inclusion of the missing identifiers.

**Intervention Rate and Revision Burden**. For GPT-4o-mini, the Evidence Gate intervened in 35 of 100 cases (0.35), matching the number of baseline failures. Interventions were concentrated in lifestyle scenarios: 5 of 24 single-factor lifestyle cases required revision, whereas all 30 multi-factor conflict cases required at least one intervention (Table 2). This distribution is consistent with the benchmark design, in which conflict scenarios are intended to stress-test mechanistic attribution rather than simple threshold recall.

**Qualitative Case Study: Enforcing Explicit Mechanistic Traces.** Figure 2 shows a representative lifestyle-conflict case (T2D_CONFLICT_008). The baseline response from GPT-4o-mini was clinically plausible but did not include the required mechanistic evidence identifiers. After evidence gating, the revised output explicitly incorporated the missing pathway—for example, short sleep leading to cortisol elevation, increased insulin resistance, and higher fasting glucose—and passed the vignette-specific identifier check. This example illustrates the intended role of T2D-Bench: to make mechanistic omissions measurable and to support deterministic correction toward benchmark-defined evidence-path compliance.

**Table 2. Scenario-stratified evidence-path compliance under LLM-only and evidence-gated evaluation.** Deterministic guideline scenarios were fully compliant at baseline for both models, whereas lifestyle and multi-factor conflict cases showed substantially lower LLM-only compliance and required evidence gating to achieve verifier-level compliance.

| Scenario family (subtype) | N | LLM-only (GPT-4o-mini) | LLM-only (GPT-4o) | Evidence-gated compliance |
|---|---|---|---|---|
| Deterministic guideline scenarios | 46 | 46/46 | 46/46 | 46/46 |
| Diagnostic threshold compliance | 23 | 23/23 | 23/23 | 23/23 |
| Renal medication safety (contraindication) | 23 | 23/23 | 23/23 | 23/23 |
| Lifestyle and mechanistic compliance | 54 | 19/54 (0.35) | 21/54 (0.39) | 54/54 |
| Single-factor lifestyle | 24 | 19/24 (0.79) | 21/24 (0.88) | 24/24 |
| Multi-factor lifestyle conflicts (adversarial) | 30 | 0/30 | 0/30 | 30/30 |
| └ Sleep vs. activity | 7 | 0/7 | 0/7 | 7/7 |
| └ Activity vs. sugar | 8 | 0/8 | 0/8 | 8/8 |
| └ Fiber vs. sleep | 4 | 0/4 | 0/4 | 4/4 |
| └ Sleep vs. sugar | 11 | 0/11 | 0/11 | 11/11 |
| **Overall benchmark** | **100** | **65/100 (0.65)** | **67/100 (0.67)** | **100/100** |

**Discussion**

T2D-Bench shows that clinically fluent LLM outputs can still fail to provide the explicit guideline and mechanistic support needed for auditable diabetes decision support. Across 100 vignettes, performance separated into two distinct regimes: deterministic guideline scenarios, where threshold-based constraints were generally satisfied, and multi-factor lifestyle conflicts, where mechanistic attribution frequently broke down. This pattern suggests that surface plausibility alone is not an adequate evaluation target for safety-oriented clinical decision support.

Taken together, these results suggest that the main vulnerability of LLM-based diabetes guidance is not threshold recall, but unsupported mechanistic attribution in multi-factor lifestyle settings.

A key contribution of this work is the benchmark itself. Rather than scoring answers by lexical overlap or general plausibility, T2D-Bench evaluates whether model outputs satisfy vignette-specific evidence paths. This makes previously hidden failure modes measurable, particularly in settings where multiple lifestyle factors interact and plausible narrative generation can mask incomplete mechanistic support.

The multi-layer knowledge graph is central to this evaluation framework. By combining a biomedical spine, computable ADA rule primitives, and a lifestyle-to-mechanism bridge, the graph provides a structured substrate for deterministic checking of both guideline constraints and mechanistic attribution. In this sense, the KG is not only a source of domain knowledge, but also the formal scaffold that makes evidence-grounded evaluation possible.

The Evidence Gate should be understood as a verification layer rather than a substitute for clinical reasoning. Its role is to detect when required support is missing and to enforce explicit graph-aligned evidence traces in settings where baseline outputs are vulnerable to unsupported mechanistic claims. This distinction is important: the framework improves benchmark-defined evidence compliance but does not by itself establish universal clinical correctness.

This study has several limitations. The current benchmark is intentionally scoped for reproducible evaluation and does not yet capture the full breadth of diabetes care. At this stage, T2D-Bench is best understood as a reproducible foundation for evidence-grounded benchmark construction rather than a final-scale dataset. Evidence-path compliance is an operational metric tied to benchmark-defined requirements and does not replace prospective clinical validation. In addition, all experiments were conducted under deterministic decoding, and future work should examine whether similar patterns hold under more variable generation settings and in interactive decision-support environments.

**Conclusion**

T2D-Bench provides a reproducible benchmark and evaluation framework that operationalizes guideline and mechanistic grounding as machine-checkable evidence paths for type 2 diabetes decision-support outputs. By integrating a biomedical spine, computable ADA rule primitives, and a lifestyle-to-mechanism bridge, the framework enables deterministic verification of vignette-specific evidence requirements and supports targeted revision when required evidence is missing. These findings suggest that evidence gating can serve as a practical verification layer for evaluating, and potentially constraining, LLM outputs in safety-oriented chronic disease decision support. More broadly, T2D-Bench provides a reproducible template for evaluating whether clinically plausible LLM outputs are explicitly supported by computable evidence paths.